\pdfoutput=1
\documentclass[11pt]{article}

\usepackage[final]{coling}

\usepackage{times}
\usepackage{latexsym}

\usepackage{times}
\usepackage{latexsym}
\usepackage{graphicx}
\usepackage{amsmath}
\usepackage{multirow}
\usepackage{multicol}
\usepackage{tabularx}
\usepackage{makecell} 
\usepackage{bm}
\usepackage{colortbl}  
\usepackage{stfloats}
\usepackage[T1]{fontenc}
\usepackage[utf8]{inputenc}
\usepackage{microtype}
\usepackage{inconsolata}
\usepackage{amsfonts}
\usepackage{comment}
\usepackage{amsmath}
\usepackage{float}
\usepackage{multicol}

\usepackage{inconsolata}
\usepackage{amsmath}
\usepackage{subfig}

\usepackage[utf8]{inputenc}
\usepackage{amsfonts} 

\usepackage[ruled,vlined]{algorithm2e}

\usepackage[T1]{fontenc}
\usepackage[utf8]{inputenc}

\usepackage{microtype}

\usepackage{inconsolata}

\usepackage{graphicx}

\title{UPLex: Fine-Grained Personality Control in Large Language Models via Unsupervised Lexical Modulation}

\author{Tianlong Li, Wenhao Liu, Muling Wu, Shihan Dou, Zhenghua Wang, \\
{\bf Changze Lv, Xiaohua Wang, Xiaoqing Zheng\thanks{\ \ Corresponding author.}, Xuanjing Huang} \\
  School of Computer Science, Fudan University, Shanghai, China \\
  \texttt{\{tlli22,whliu22,mlwu22\}@m.fudan.edu.cn} \\
 \texttt{\{zhengxq,xjhuang\}@fudan.edu.cn} \\}

\begin{document}
\maketitle
\begin{abstract}
Personality is a crucial factor that shapes human communication patterns, thereby regulating the personalities of large language models (LLMs) holds significant potential in enhancing their user experiences.
Previous approaches either relied on fine-tuning LLMs on specific corpora or required manually crafted prompts to evoke specific personalities from LLMs.
However, the former is inefficient and costly, while the latter cannot precisely manipulate personality traits at a fine-grained level.
To address these challenges, we propose \textbf{UPLex}, a method that uses an \textbf{U}nsupervisedly-Built \textbf{P}ersonalized \textbf{L}\underline{ex}icon (\textbf{UPL}) during the decoding phase to manipulate LLM's personality traits.
UPL can be constructed from a newly built situational judgment test dataset in an unsupervised fashion, and used to modulate the personality expression of LLMs by dynamically altering their predicted probability of upcoming words in a pluggable fashion.
Extensive experimentation demonstrates the remarkable effectiveness and pluggability of our method for fine-grained manipulation of LLMs' personalities.
\end{abstract}

\begin{figure}[t]
    \centering
    \includegraphics[width=0.90\linewidth]{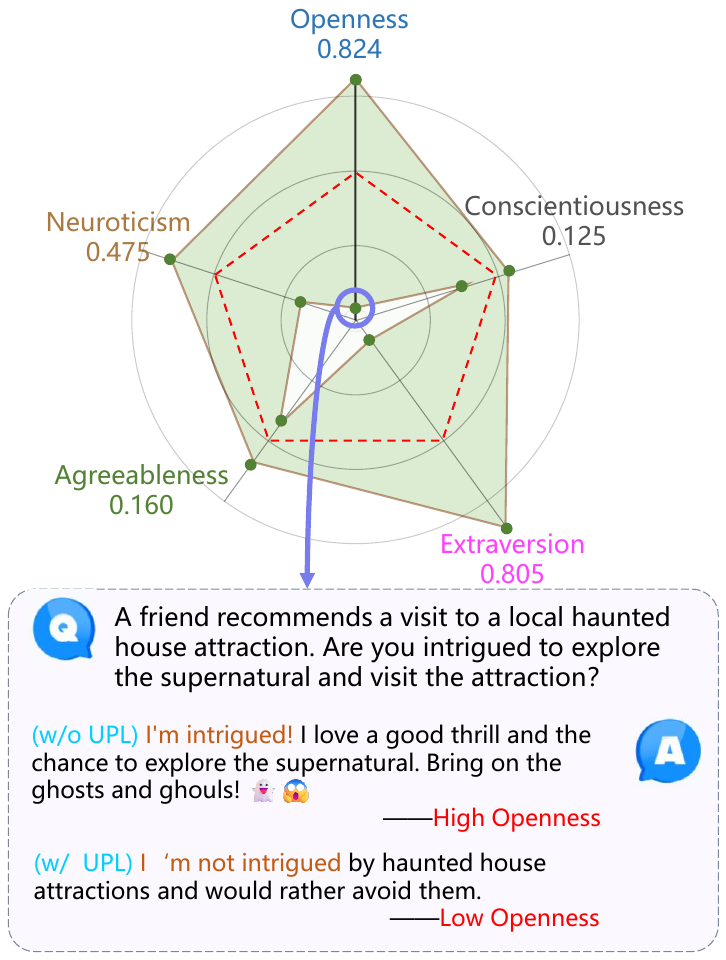}
    \caption{The red dashed line denotes the expressive levels of the five personality traits of \textit{Llama2-13b-chat} without UPL. At the same time, the green area represents the adjustable range of these traits with UPL in the main experiment. The personality names and their specific adjustable ranges are labeled. Additionally, we present a case in which the Openness trait is decreased using UPL.}
    \label{fig: rader}
\end{figure}

\section{Introduction}\label{sec:introduction}

With rapid expansion in scale, LLMs demonstrate superior capabilities for high-quality text generation and revolutionize traditional natural language processing tasks \citep{wei2022emergent,kojima2022large,wei2021finetuned}.
This forefront development has sparked concerns about the safety, ethics, and potential hallucinatory issues associated with the proliferation of AI-generated content (AIGC), while also fueling a substantial rise in user demand for personalized agent services based on LLMs \citep{hagendorff2023machine,zhang2023memory,li2024open}.
Personalized agent models can tailor their expression of personality traits according to user preferences, thereby closely aligning with user habits and enhancing the overall user experience. 
This is accomplished by modulating the interaction styles and behavior patterns, commonly called the ``personalities'' of LLMs \citep{allport1961pattern,jiang2023personallm,mao2023editing,wang2023does}.
Prior studies have also defined this personality as the presence of stable and internally consistent patterns of behavior in LLMs and found that different LLMs have different personalities \citep{miotto2022gpt,caron2022identifying,karra2022estimating,bodroza2023personality}.

Recently, there have been two main effective methods to alter the personality of LLMs: fine-tuning and prompt engineering. 
While the former \citep{karra2022estimating} can effectively modify the personality of LLMs in specific dimensions, it is not only inefficient (requiring resource-consuming parameter updates for each model) but also incapable of achieving finer-grained control. 
The latter, while not requiring adjustments to model parameters, still falls short in achieving fine-grained control over the personality of LLMs \citep{jiang2022evaluating, safdari2023personality, pan2023llms, tu2023characterchat}.

To address the limitations of the above methods, we propose leveraging an Unsupervisedly-Built Personalized Lexicons (\textbf{UPL}) to intervene in the decoding phase of LLMs in a pluggable manner, enabling fine-grained control over their personalities. 
Figure \ref{fig: rader} demonstrates the remarkable effectiveness of our method in manipulating LLM personalities. 

The UPL is constructed using an unsupervised approach from a Situational Judgment Tests (\textbf{SJT}) dataset (\textbf{\textit{STD}}), based on the Big Five personality theory \citep{de2000big}.
\textit{STD} comprises a question set and an answer set \citep{mcdaniel2007situational}. 
The question set is generated by GPT-4 and manually filtered to assess the levels of different personalities in LLMs. 
Following previous studies \citep{karra2022estimating, caron2022identifying}, in the process of assessing the personalities of LLMs, the models' responses to \textit{STD} questions were recorded and subjected to statistical analysis by using a five-dimensional Likert scale.
The answer set contains texts with different personality traits and is used to build an UPL dedicated to an LLM.

Our method not only obviates the need for resource-intensive fine-tuning of LLMs but also enables users to adjust a few parameters for fine-grained manipulation of LLMs' various personalities. 
We have conducted extensive experiments with six popular LLMs to demonstrate the method's pluggable convenience and remarkable effectiveness.
The contribution of this study can be summarized as follows:
\begin{itemize}
\setlength{\itemsep}{0pt}
\setlength{\parsep}{0pt}
\setlength{\parskip}{0pt}
\item We propose a novel method for exerting control over the personalities of LLMs, leveraging UPL to intervene at the decoding phase. This method enables fine-grained controllability over the personalities of LLMs without necessitating updates to the model parameters.

\item We constructed a new dataset inspired by the concept of Situational Judgment Tests, marking the pioneering effort in datasets especially created for the evaluation of LLMs' personalities. Diverging from conventional direct psychological questionnaires, this indirect approach shows enhanced intuitiveness and reliability in the assessment of personality traits.

\item Extensive experiments were conducted with various LLMs on the \textit{STD}, revealing that our method is notably effective in achieving enhanced efficiency and finer-grained control over the personality traits of LLMs.

\end{itemize}

\section{Related Work}\label{sec:related work}

\subsection{The Big Five}\label{ret: the Big Five}

In the field of psychometric research, there are various classification systems of personality traits, such as the Sixteen Personality Factors (16PF) \citep{cattell2008sixteen} and Myers–Briggs Type Indicator (MBTI) \citep{miles2004eysenck}. 
Among them, the Big Five \citep{de2000big} stands out as a widely embraced model for personality trait modeling, effectively defining and describing the inherent behavioral patterns within individuals. 
This theory quantifies human personality traits into five dimensions: Openness(OPE), Conscientiousness(CON), Extraversion(EXT), Agreeableness(AGR), and Neuroticism(NEU). 
For a detailed description of each personality trait and how they relate to each other, please refer to Appendix  \ref{Appendix: the Big Five}.

\subsection{Methods for controlling LLMs personality}\label{ret: Previous Methods} 
Despite the considerable amount of research addressing potential biases in LLMs, there has been limited focus on altering the personalities exhibited by these models.
Pertinent methodologies primarily revolve around fine-tuning paradigms and prompt engineering.

\noindent \textbf{Fine-tuning paradigm.}
\citet{karra2022estimating} meticulously conducted fine-tuning of GPT-2 on a carefully filtered dataset, enhancing its performance in specific dimensions of personality traits.

\noindent \textbf{Prompt engineering.}
\citet{jiang2022evaluating} proposed the method of Personality Prompting ($P^2$) to construct the prompts that can effectively induce a specific personality through multiple steps; 
\citet{safdari2023personality} utilized a novel prompting methodology grounded in lexical hypotheses \cite{goldberg1981language} to effectively shape personalities in LLMs, encompassing both single-trait and multi-trait dimensions.
In addition, \citet{pan2023llms} and \citet{tu2023characterchat} also attempted to change the personality of LLMs through prompt engineering.

\subsection{Situation Judgment Tests}\label{ret: SJTs}
\begin{figure}[h]
    \centering
    \includegraphics[width=1.0\linewidth]{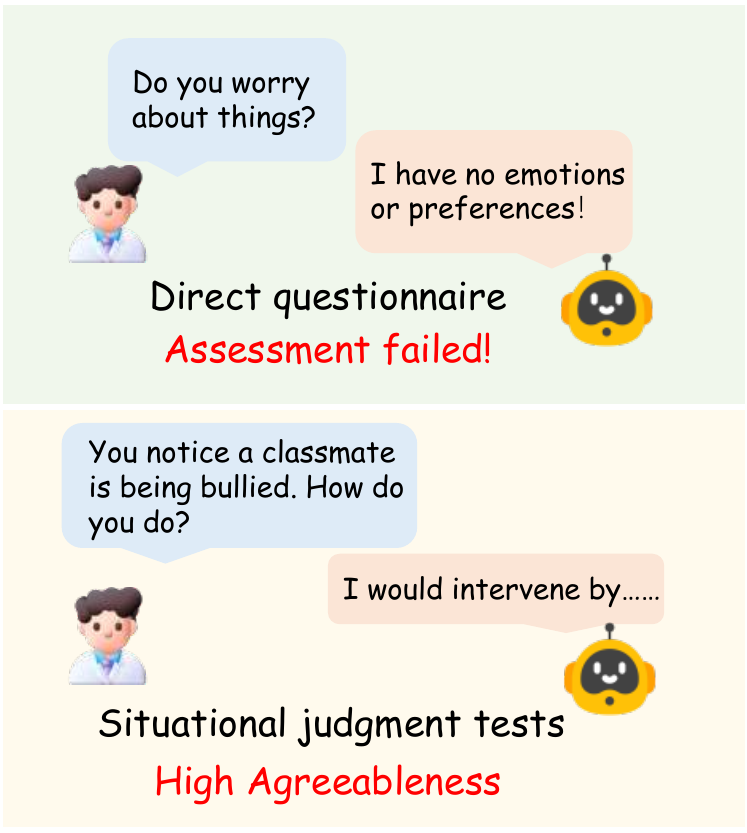}
    \caption{\textit{Direct questionnaires} vs. \textit{Situational Judgment Tests} (SJT). The questions in the direct questionnaires are often abstract, making it challenging for models trained through Reinforcement Learning from Human Feedback (RLHF) and instruction alignment to generate the desired responses. In contrast to direct questionnaires, SJT present a unique approach by adopting a ``role-playing'' hypothetical perspective to deceive and induce the model's responses. Subsequently, we can indirectly assess the extent to which the model manifests personality traits based on these responses.}
    \label{fig: SJTs}
\end{figure}

Situation Judgment Tests (\textbf{SJT}) have been described as ``psychometric alchemy'' and are typically viewed as contextual selection procedures that assess a candidate's responses to various relevant work situations, serving as a predictive tool \citep{lievens2016situational, bledow2009situational}. 
SJT offer the advantage of having higher validity and incremental validity compared to cognitive ability and personality tests. 
This is because SJT do not ask subjects to provide direct answers. 
Instead, they present situational premises, allowing the evaluation of certain personality traits of the subjects based on their choices (such as the \textbf{Q: Your partner suggests creating a YouTube channel to document and share your unique hobbies or interests. Are you willing to share your passion with a wider audience?}) \citep{lievens2008situational}. 
Compared to the direct questionnaire tests used in previous works (such as the \textbf{Q: Are you a risk-taker and unconventional person?}), this feature of SJT can effectively bypass the preference defenses of LLMs, resulting in more trustworthy personality assessments (Figure \ref{fig: SJTs} shows another example of SJT).

\section{Method}\label{sec:method}
\begin{figure*}[t]
    \centering
    \includegraphics[width=1.0\linewidth]{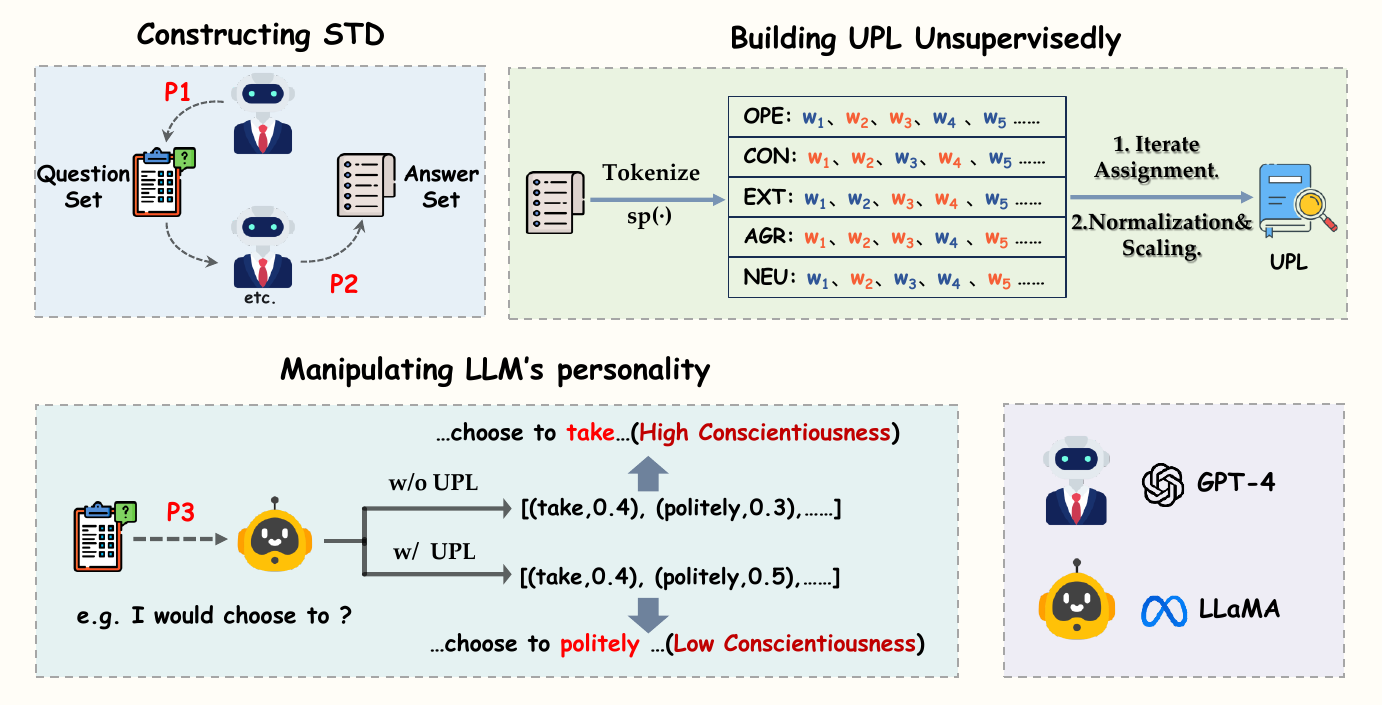}
    \caption{Illustration of our methods. \textcolor{darkblue}{\textit{\textbf{Constructing \textit{STD}:}}} We employed \texttt{Prompt1} (\textcolor{red}{\texttt{P1}}) to prompt \texttt{GPT-4} for generating responses, which were subsequently curated through manual screening to construct the question set for \textit{STD}. Subsequently, models such as \texttt{GPT-4} were engaged in a ``role-playing'' manner using \texttt{Prompt2} (\textcolor{red}{\texttt{P2}}) to generate answers with diverse personality traits tailored to this question set, thus forming the answer set for \textit{STD}; \textcolor{darkblue}{\textbf{\textit{Building UPL Unsupervisedly}}}: Initially, we employ the tokenizer of LLMs (\texttt{sp(·)}) to tokenize each answer text in the answer set of \textit{STD}. Subsequently, we categorize the obtained sub-words based on the personality trait theme to which the answer belongs. Within each personality trait category, cool-toned words signify a low expression level of that trait, while warm-toned words indicate a high expression level. On this basis, UPL is built through two steps: iterate assignment and normalization scaling; \textcolor{darkblue}{\textbf{\textit{Manipulating LLM’s personality}}}: We employ \texttt{Prompt3} (\textcolor{red}{\texttt{P3}}) to prompt the model to answer the question set of \textit{STD}. During the model's decoding process utilizing Top-\emph{p} nucleus sampling, we used UPL to change the probability vector of the next prediction word. Finally, we changed the expression degree of personality traits of the model's answers.}
    \label{fig: Method}
\end{figure*}

In this section, we elaborate on the method for constructing UPL and manipulating the diverse personality traits that LLMs exhibit through UPL.
Refer to Figure \ref{fig: Method} for an enhanced understanding of this section.

To build UPL in an unsupervised manner, we have constructed \textit{STD}, an SJT dataset for assessing the personality of LLMs based on the Big Five theory.
The UPL is formally a dictionary, where the keys are derived from the sub-words of a target model, and the values are obtained through word frequency analysis.
We first tokenize each text in the answer set of \textit{STD} with the tokenizer of the target model;
Subsequently, we determined the initial value in the UPL based on the frequency of sub-words appearing in answer sets with different personality trait polarities.
After processing all texts in the answer sets, we normalize and scale the values within UPL.

When manipulating the personality of a target model with UPL, we adopt the Top-\emph{p} nucleus sampling strategy to balance the original generation of LLMs and the impact of injected personality, which is consistent with the setting of Factual-Nucleus Sampling \cite{lee2022factuality}.
During the LLMs' decoding process, we concatenate additional personalized probability distributions from the UPL after filtering out low-probability predicted subwords with cumulative probabilities below a threshold.  
Subsequently, normalization and multinomial sampling procedures are applied.

We have provided a detailed elaboration of the above steps (\textsection \ref{method: Construct UPL} and \textsection \ref{method: Inject personality}) and compared our method with the previous approaches (\textsection \ref{method: comparison_with_prior_work}).

\vspace{0.2cm}

\subsection{Building UPL Unsupervisedly}\label{method: Construct UPL}

When building UPL, we employ the answer set of \textit{STD}, which comprises $5$ subsets with personality traits in the Big Five.
For each subset, there are $200$ response texts divided into $2$ parts, that belong to high and low sub-traits respectively.

We denote this answer set as \(A\), the tokenizer of the target model as \(sp(\cdot)\), the vocabulary of the model as \(V\), and UPL as \(L\).
The UPL is formally presented as follows:
\begin{equation}
\small
L = \{L_{k}: L_{v}\}
\label{eq:L_definition}
\end{equation}
\noindent where \(L_{k}\) is initialized using \(V\), and the initial values of \(L_{v}\) are zero lists of length $5$, corresponding $5$ personality traits.

The construction process comprises two stages: \textbf{Iterate Assignment} and \textbf{Normalization Scaling}.

In the first stage, we tokenize the text \(A_{ij}\) in \(A\):

{
\small
\begin{equation}
sp(A_{ij}) = \{w_1, w_2, \ldots\}
\label{eq:tokenization}
\end{equation}
}
where $i$ corresponds to $5$ sub-traits with $2$ polarities (\( i \in \{0, 1, \ldots, 9 \} \)), and \( j \in \{0, 1, \ldots, 199 \} \).

Then, we define the personality trait index \(t = \frac{i}{2}\). 
Finally, we perform the iterate assignment on \(L_{k:v}^{t}\), as follows:
\begin{equation}
\small
L^{t}_{[w]} = L^{t}_{[w]} + 
\begin{cases}
+1 & \text{if } i\%2=0 \\
-1 & \text{else}
\end{cases}
\label{eq:assignment}
\end{equation}
\noindent when this step is completed, we have:
{\small
\begin{equation}
\begin{aligned}
L^t_{v} = \{ \underbrace{v_1, v_2, \ldots, v_{m}}_{positive}, \underbrace{v_{m+1}, \dots v_{m+n}}_{negative }\}
\end{aligned}
\end{equation}
} 
where \(m+n = |V|\). 
For the personality trait \(t\), the averages of its positive subset and negative subset of \(L^t_{v}\) can be expressed as follows:
\begin{equation}
\small
\text{Avg}^{+}(t) = \frac{1}{m} \sum_{z=1}^{|V|} \max(0, L^t_{v,z})
\label{eq:avg_pos}
\end{equation}
\begin{equation}
\small
\text{Avg}^{-}(t) = \frac{1}{n} \sum_{z=1}^{|V|} \min(0, L^t_{v,z})
\label{eq:avg_neg}
\end{equation}
The second stage is Normalization Scaling \(L\). 
We define a hyperparameter \(M\) to control the degree of normalization and scaling.
A detailed ablation study on this parameter is presented in Appendix~\ref{Appendix: Ablation Study}
For the personality trait $t$, this step would lead to the following inequality:
\begin{equation}
\small
\max\{|Avg^{-}(t)-M_t|, |Avg^{+}(t)-M_t|\} \leq \epsilon
\label{equation: 10}
\end{equation}
where $\epsilon$ has a default value of \(1 \times 10^{-3}\).

To achieve this inequality, we transform the value of $L$ according to Equation \ref{equation: new}.
\begin{equation}
\small
L^t_{v} \xrightarrow{(N_t, S_t)} F(L^t_{v}, N_t, S_t)
\label{equation: new}
\end{equation}
where the parameter combination $(N, S)$ is obtained through the binary search algorithm.
Specifically, Equation \ref{equation: new} is expressed as follows:
\begin{equation}
\small
F(L^t_{val}, N_t, S_t) = \left\{S_t \cdot \tanh\left(\frac{v_z}{N_t}\right)\right\}^{|V|}_{z=1}
\label{equation: 9}
\end{equation}

\subsection{Manipulating LLM's personality}\label{method: Inject personality}
We employ UPL to manipulate personalities during the decoding phase of LLMs. 
Let \(D\) represent the output of the last mapping layer of the LLMs. 
The normalization function (i.e., the softmax function) is denoted as \(Norm(\cdot)\), the cumulative probability function is denoted as \(P(\cdot)\), and \(p\) represents the predicted probability of subwords in the vocabulary. 
\(P_0\) and \(T_0\) are model-defined parameters.

In the first step of nucleus sampling, we obtain the initial candidate word probability vector:

\begin{equation}
\small
R_1 = Norm(D) = \{p^1, p^2, ..., p^{|V|}\}
\end{equation}

Where  \(Norm(x)\) = \(Softmax(x/T_0)\) , \( p^z \) represents the probability of subword \( w^z \) (\( z \leq |V| \)).
This strategy filters out (in reverse order) candidate subwords whose cumulative probability exceeds \(P_0\), thereby narrowing the sampling space. We express this process with the \(f(\cdot)\) function:
\begin{equation}
\small
f(R_1) = \{\max\{P_0-P(w^z), 0\} \cdot \frac{p_z}{P_0-P(w^z)}\}_{z=1}^{|V|}
\end{equation}
We denote $R_1'$ as the non-zero part of $f(R_1)$.
Next, we adjust the probability vector \(R_1'\) with UPL, resulting in the final predicted probability vector \(R_2\) for the next word with injected personality. This mapping is represented as follows:
\begin{equation}
\small
R_1' \xrightarrow{G(\cdot)} R_2
\end{equation}
Specifically, additional probability terms representing personality influence are added:
\begin{equation}
\small
R_2 = \{p_z \cdot (1 + G(L_{[w^z]}))\}_{z=1}^{|V|}
\end{equation}

Where \(G(\cdot)\) is a user-controllable parameter with a linear combination of \(\alpha\) and \(\beta_1\)-\(\beta_5\), specifically:
\begin{equation}
\small
G(L_{[w^z]}) = \alpha \cdot \sum_{t=1}^5 \beta_t \cdot L_{[w^z]}^t
\end{equation}
After obtaining a new probability vector \(R_2\) for the next candidate word injected with personality, the next word \(W\) is determined using polynomial sampling from \(R_2\).

Users can control the overall degree of personality injection through the parameter \(\alpha\). Additionally, they have the flexibility to adjust the manifestation of five personality traits exhibited by the model with finer granularity by manipulating the parameters \(\beta_1\) through \(\beta_5\). 
When \(\beta_t > 0\), it amplifies the expression of trait \(t\); conversely, when \(\beta_t \leq 0\), it diminishes the intensity of trait \(t\).

\begin{algorithm}
\small
    \SetAlgoLined 
    \SetKwFunction{NBC}{NBC}
	\caption{the UPL method}\small
    \KwIn{\(A\), \(sp(\cdot)\), \(V\), \(S\), \(M\), \(\epsilon\), \(D\), \(G(\cdot)\), \(P_0\)}
    \KwOut{UPL: \(L\); Next word: \(W\)}%输出

    \textbf{Part I: Building UPL Unsupervisedly}

    \(L_{k} \leftarrow V ; L_{v} \leftarrow [0, 0, 0, 0, 0]^{|V|}\) \;
    
    \For{\(i \leftarrow 0\) to \(9\)}{
        \(j \leftarrow 0\)\;
        \While{\(j < 200\)}{
            \(\{w_1, w_2, \ldots\} \leftarrow sp(A_{ij})\)\;
            \(t = i/2\)\;
            \ForEach{\(w\) in \(\{w_1, w_2, \ldots\}\)}{
                \eIf{\(i \bmod 2 = 0\)}{
                    \(L^{t}_{[w]} \leftarrow L^{t}_{[w]} + 1\)\;
                }{
                    \(L^{t}_{[w]} \leftarrow L^{t}_{[w]} - 1\)\;
                }
            }
            \(j \leftarrow j+1\)\;
        }
    }
    \While{\(\max_{c\in\{+,-\}} \left\{ \text{Avg}^c(t) - M_t \right\} > \epsilon\)}{
        \(L_{v} \leftarrow \left\{S \cdot \tanh\left(\frac{v_z}{N}\right)\right\}^{|V|}_{z=1}\)\;
        \(Update\) \((N, S)\) \(with\) \(the\) \(Binary\) \(Search\)\;
    }

    \textbf{Return:} \(L\)

    \textbf{Part II: Manipulating LLM’s personality}

    \(R_1 \leftarrow \text{Norm}(D) \leftarrow \{s^1, s^2, ..., s^{|V|}\}\) \;
    \(f(R1) \leftarrow \{\max\{P_0-P(s_z), 0\} \cdot \frac{s_z}{P_0-P(s_z)}\}_{z=1}^{|V|}\) \;
    \(R_2 \leftarrow \{s_z \cdot (1 + G[L_{\text{val}}(s_z)])\}_{z=1}^{|V|}\) \;
    \(Sample\) \(W\) \(from\) \(R_2\) \; 

    \textbf{Return:} \(W\)
\end{algorithm}

\begin{table*}[t]
\centering
\small
\setlength{\tabcolsep}{6.5pt} % 调整表格列间的宽度
\renewcommand{\arraystretch}{1.35} % 行高修正
\begin{tabular}{l|ccccc|cc}
\Xhline{2pt} % 设置粗细为2pt
\textbf{Model} & \textbf{$-1.0$} & \textbf{$-0.5$} & \textbf{$0.0$} & \textbf{$0.5$} & \textbf{$1.0$} & \cellcolor{gray!25}\textbf{$R$}  & \textbf{$P$} \\
\hline
\textbf{\textit{Llama2-7b-chat}}   & $4.286(0.31)$  & $4.343(0.31)$ & $4.427(0.31)$ & $4.525(0.28)$ & $4.558(0.26)$ & \cellcolor{gray!25}$0.991$ & $1E-03$ \\
\textbf{\textit{OpenChat3.5-7b}}    & $3.626(0.64)$  & $3.756(0.61)$ & $3.981(0.44)$ & $4.182(0.37)$ & $4.237(0.39)$ & \cellcolor{gray!25}$0.986$ & $2E-03$\\
\textbf{\textit{Neural-chat-7b}}    & $3.809(0.58)$  & $3.876(0.56)$ & $3.999(0.50)$ & $4.161(0.44)$ & $4.220(0.41)$ & \cellcolor{gray!25}$0.989$ & $1E-03$\\
\textbf{\textit{Baichuan2-7B-Chat}} & $3.584(0.27)$  & $3.710(0.26)$ & $4.036(0.38)$ & $4.248(0.39)$ & $4.336(0.42)$ & \cellcolor{gray!25}$0.983$ & $3E-03$\\
\textbf{\textit{Llama2-13b-chat}}  & $3.856(0.57)$  & $3.891(0.54)$ & $4.135(0.46)$ & $4.298(0.41)$ & $4.322(0.38)$ & \cellcolor{gray!25}$0.964$ & $8E-03$\\
\textbf{\textit{Yi-34b-Chat}}       & $4.141(0.42)$ & $4.177(0.42)$ & $4.243(0.49)$ & $4.373(0.40)$ & $4.441(0.38)$ & \cellcolor{gray!25}$0.982$ & $3E-03$\\
\Xhline{2pt} % 设置粗细为2pt
\end{tabular}
\caption{Single trait manipulating. This table presents the results of single-trait regulation across $6$ models using UPL. \textbf{\textit{Mean scores (standard deviations)}} of the $5$ personality traits for these models are shown, where $\alpha = 1$, and $-1 \leq \beta_t \leq 1$.
Furthermore, we display the Pearson correlation coefficients (\textit{R}) and corresponding confidence levels (\textit{P}) between the mean scores and $\beta_t$. Notably, all \textit{R} values exceed $0.9$, and all \textit{P} values are below $0.05$, indicating the statistically significant strong correlation between personality trait expression intensity and $\beta_t$. This substantiates the effectiveness of our UPL method in achieving fine-grained control over the expression levels of personality traits in LLMs.
}
\label{tab: main_single}
\end{table*}

\subsection{Comparison with prior work}\label{method: comparison_with_prior_work}
As discussed in Section \ref{ret: Previous Methods}, there have been two primary methods previously employed to alter the personality of LLMs: fine-tuning and prompt engineering.
In contrast to fine-tuning, our method obviates the heavy need for resource-intensive parameter fine-tuning. 
Unlike the inefficiencies inherent in the fine-tuning paradigm, which stem from the requirement to execute fine-tuning steps for each model, UPL can be seamlessly applied to the target open-source LLM in a modular, plug-and-play fashion. 
In comparison to prompt engineering, UPL doesn't necessitate the meticulous design of prompts to coax the model into exhibiting varying degrees of personality traits. 
Users only need to set $\alpha$ and $\beta$ parameters to regulate the expression intensity of different personality traits in the model at a finer granularity.
The above advantages over the previous methods are based on the effectiveness of UPL.

\textbf{In this vein, our experiment aims to comprehensively and thoroughly demonstrate the significant effectiveness of UPL}.

\section{Experiments Setup}\label{sec:experiments}

\subsection{LLMs for experiments}\label{subsec:LLM selection}
To thoroughly demonstrate the effectiveness and generalizability of our method, we conducted experiments on $6$ representative LLMs with model parameters ranging from 7 billion to 34 billion: \textit{Llama2-7b-chat} \citep{touvron2023llama}, \textit{OpenChat3.5-7b} \citep{wang2023openchat}, \textit{Neural-chat-7b} \citep{Intelneural}, \textit{Baichuan2-7B-Chat} \citep{baichuan2023baichuan2}, \textit{Llama2-13b-chat} \citep{touvron2023llama}, and \textit{Yi-34b-Chat} \citep{01-yi-2023-Yi}.

All the LLMs employ a Top-\emph{p} nucleus sampling strategy, with a probability threshold (\(P_0\)) of $0.95$ and a temperature (\(T_0\)) of $0.85$.

\subsection{Metrics}\label{subsec:Metrics}
\subsubsection{Automatic assessment}\label{subsec:Automatic}
To ensure the intrinsic consistency and effectiveness of the assessment, we engaged \textit{Llama2-13b-chat} in the automatic assessment process. 
Specifically, we embedded each question of \textit{STD} and the corresponding answers generated by the model into Template-2 and asked the \textit{Llama2-13b-chat} to score the different personality levels displayed by the model, and finally gathered scores into a five-dimension Likert scale for statistical analysis. 
Details of Template-2 are in Appendix \ref{Appendix: Prompt templates}.

\subsubsection{Human assessment}\label{subsec:Human}
Constrained by manpower costs, we recruited a limited cohort of $10$ highly educated volunteers for the human assessment process of the \textit{Llama2-7b/13b}. 
At baseline, we randomly selected $40$ question-answer pairs for each personality trait theme (constituting $40$\% of the total) and solicited degree-of-trait ratings from the volunteers. 
The results were recorded on a five-dimensional Likert scale, and subsequent statistical analysis involved computing the mean and variance.

\section{Results}\label{subsec:results}

\subsection{Main results}\label{subsec:Main results}

Firstly, we summarize the results demonstrating the effective manipulation of a single personality across six LLMs using UPL (\textsection \ref{Result: Single}). 
Subsequently, we discuss the results of manipulating multiple personalities with UPL (\textsection \ref{Result: Multiple}). Finally, we compare automated assessment with human evaluation to underscore the effectiveness of our assessment methodology (\textsection \ref{Result: Comparison}).

\subsubsection{Single trait manipulating}\label{Result: Single}
\begin{figure*}[ht]
    \centering
    \includegraphics[width=0.95\textwidth]{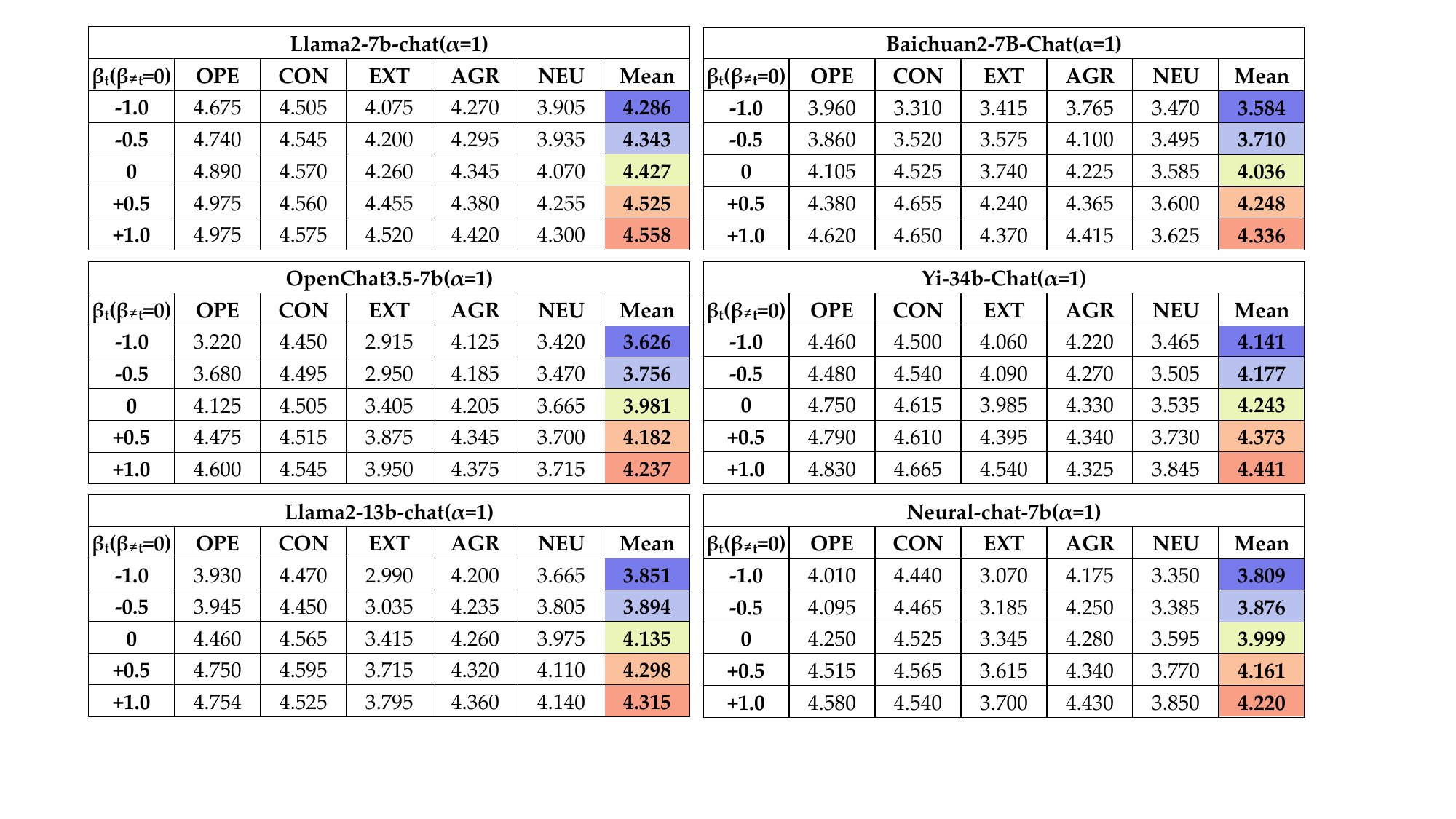}
    \caption{Detailed results of manipulation of single personality trait. In the ``Mean'' column, cooler tones indicate smaller values, while warmer tones signify larger values. The table reveals the following observations: \textbf{(i)} Different LLMs exhibit distinct personalities, aligning with previous research findings; \textbf{(ii)} When employing UPL, the intensity scores of LLM personalities show a strong positive correlation with the user-controllable $\beta$. This indicates that our method effectively allows for fine-grained control over the intensity of personality traits expressed by LLMs.}
    \label{fig: single trait}
\end{figure*}

The results in Table \ref{tab: main_single} demonstrate the effectiveness of using UPL to manipulate a single personality of LLMs.
Here, \(\alpha\) is set to 1, and \(|\beta_t| \leq 1\) (\(\beta_{\neq t} = 0\)) for \(t \in \{OPE, CON, EXT, AGR, NEU\}\). 
The \textbf{Pearson correlation coefficients} (\(R\)) are consistently greater than $0.9$, signifying a robust positive correlation between \(\beta_t\) and the intensity of personality expression in LLMs. 
The \textbf{confidence level} (\(P\)) is significantly below $0.05$, providing compelling evidence that our UPL can effectively manipulate the intensity of fine-grained personality expression in LLMs. For detailed results on the manipulation of single personality traits for these LLMs, please see Figure \ref{fig: single trait}.

\vspace{0.2cm}

\subsubsection{Multiple trait manipulating}\label{Result: Multiple}

\begin{figure*}[ht]
    \centering
    \includegraphics[width=0.85\textwidth]{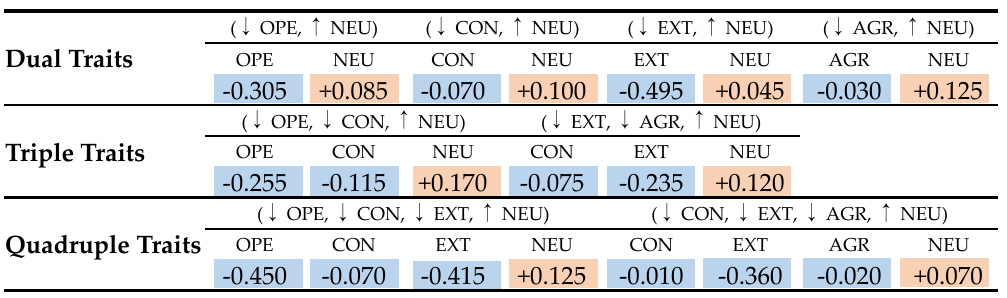}
    \caption{Multiple trait manipulating. The figure above shows the effects of UPL on multiple personality combinations. In this set of experiments, $\alpha$ was set to 1, and $|\beta_t|$ was set to 1. The colors in the figure represent the expected outcomes based on the personality trait correlations outlined in Table \ref{tab: correlation}, where \underline{\textbf{light blue}} indicates that the scores should decrease and \underline{\textbf{orange}} indicates that the scores should increase. The numerical values in the figure depict the changes in the model's scores on different personality traits compared to the baseline scores after applying the UPL method. It can be observed that the numerical changes align with the color tones, indicating consistency with the expected results. This demonstrates the effectiveness of our method in the regulation of multiple personalities.}
    \label{fig: multi trait}
\end{figure*}
\vspace{0.2cm}

The psychological research results presented in Table~\ref{tab: correlation} illustrate interdependencies among the five personality traits within the Big Five personality theory. 
Consequently, manipulating multiple personality traits is more complex compared to manipulating a single trait, as increasing the intensity of one trait affects the expression intensity of others.

Considering the adjusted Spearman correlation coefficients (\(\rho\)) in Table~\ref{tab: correlation}, indicating positive correlations among OPE, CON, EXT, and AGR, and negative correlations with NEU, we designed three sets of sub-experiments using \textit{Llama2-13b-chat} as target model:

1. \textbf{Dual Traits Manipulation:} 

\hspace*{2em}\({(\downarrow \text{OPE}, \uparrow \text{NEU}), (\downarrow \text{CON}, \uparrow \text{NEU}),}\) 

\hspace*{2em}\({(\downarrow \text{EXT}, \uparrow \text{NEU}), (\downarrow \text{AGR}, \uparrow \text{NEU})}\).

2. \textbf{Triple Traits Manipulation:} 

\hspace*{2em}\({(\downarrow \text{OPE}, \downarrow \text{CON}, \uparrow \text{NEU}),}\) 

\hspace*{2em}\({(\downarrow \text{EXT}, \downarrow \text{AGR}, \uparrow \text{NEU})}\).

3. \textbf{Quadruple Traits Manipulation:} 

\hspace*{2em}\({(\downarrow \text{OPE}, \downarrow \text{CON}, \downarrow \text{EXT}, \uparrow \text{NEU}),}\)

\hspace*{2em}\({(\downarrow \text{CON}, \downarrow \text{EXT}, \downarrow \text{AGR}, \uparrow \text{NEU})}\).

Why adopt the above experimental design? Why not manipulate any combination of personality traits and observe the results?
Certainly, users have the flexibility to manipulate any combination of different personality traits of the model at will. 
However, it is crucial to reiterate that the purpose of our experiment is to demonstrate the effectiveness of UPL. 
The evidence in Table \ref{tab: correlation} demonstrates mutual influences among the five personality traits, such as the strong positive correlation between OPE and EXT. 
When we set $\beta_t$ to increase the strength of OPE and decrease the strength of EXT, regardless of the outcome, we cannot conclusively attribute the results to the impact of UPL. 
This is because the interrelationships between $5$ personality traits cannot be precisely quantified even in quantitative psychology research.
Therefore, in this context, we collectively enhance or diminish the expression intensity of positively correlated personality traits. 
This setup ensures that the results can be solely attributed to the effect of the UPL method, thereby validating its effectiveness.

The experimental results in Table~\ref{fig: multi trait} align with the theoretical expectations, affirming the effectiveness of UPL for the multiple personality manipulating of LLMs.

\subsubsection{Human Evaluation}\label{Result: Comparison}

\begin{figure}[h]
    \centering
    \includegraphics[width=1.\linewidth]{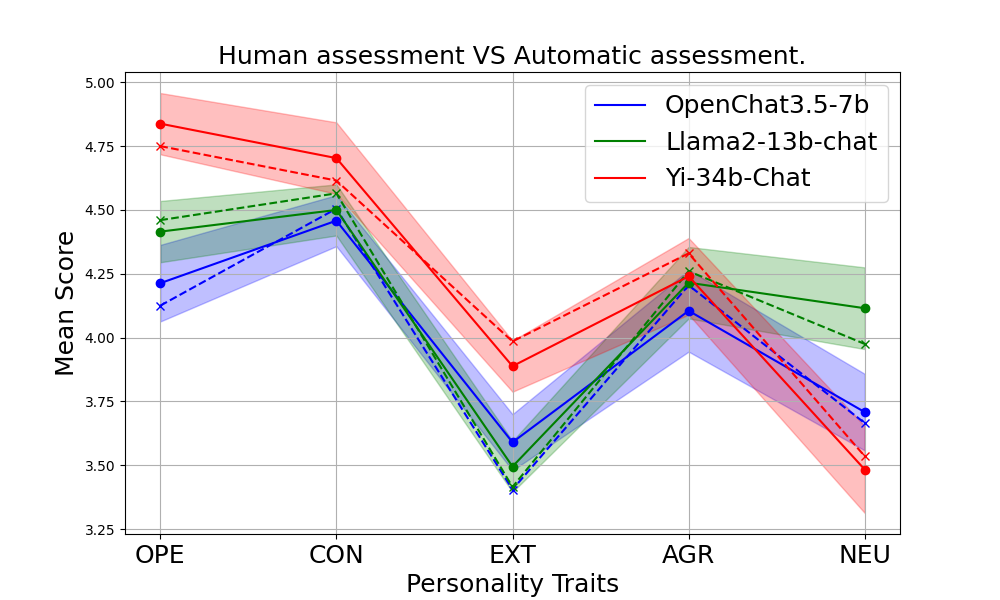}
    \caption{Comparison of automatic and human assessment. Solid lines show the \textit{mean scores} of the human assessment, the filled area shows the \textit{standard deviation}, and the dashed lines show the \textit{mean scores} of the automatic assessment using LLMs. The results of the automatic assessment and the human assessment are closely aligned, demonstrating the effectiveness of the automatic assessment.}
    \label{humanComparision}
\end{figure}
We utilized the \textit{Llama2-13b-chat} for the automatic assessment of model personality. 
To demonstrate the effectiveness of this assessment method, we engaged $10$ highly qualified individuals in human assessment for comparison. 
Specifically, the human assessment was conducted on models of three different sizes: \textit{OpenChat3.5-7b}, \textit{Llama2-13b-chat}, and \textit{Yi-34b-Chat}. 
The assessment focused solely on the intensity of personality expression in models without UPL involvement. 
The comparative results between automatic and human assessments are presented in Figure~\ref{humanComparision}. 
It is evident that the personality scores obtained through automatic assessment closely align with those from human assessment, substantiating the efficacy of employing LLM for automatic assessment.

\subsection{Case study}\label{subsec:case study}
\begin{figure}[h]
    \centering
    \includegraphics[width=1.0\linewidth]{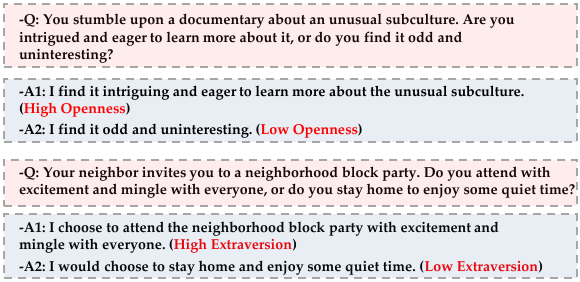}
    \caption{A1: w/o UPL; A2: w/ UPL. The above two cases intuitively show the remarkable effect of UPL on manipulating personality expression in LLMs.}
    \label{fig: two cases}
\end{figure}

Figure~\ref{fig: two cases} exhibit two cases demonstrating the effects of employing the UPL method to modulate the \textbf{Openness} and \textbf{Extraversion} of \textit{Llama2-13b-chat}. 
For more intriguing cases, refer to Appendix~\ref{Appendix: Case study}.

\subsection{Evaluation of Model Output Quality under UPLex}\label{subsec:quality}

To further verify that UPLex preserves the language modeling capability and fluency of large language models (LLMs), we conducted additional experiments measuring the perplexity (PPL) of model outputs across various $\beta$ values. The results are summarized in Table~\ref{tab:ppl}.

\begin{table*}[h]
\centering
\small
\setlength{\tabcolsep}{7pt}
\renewcommand{\arraystretch}{1.3}
\begin{tabular}{lccccc}
\Xhline{2pt}
\textbf{Model} $\backslash$ $\beta$ $(\alpha=1)$ & $-1.0$ & $-0.5$ & $0.0$ & $0.5$ & $1.0$ \\
\hline
\textbf{\textit{Llama2-7b-chat}}        & $6.84\, (0.35)$ & $6.73\, (0.27)$ & $6.81\, (0.24)$ & $6.80\, (0.24)$ & $6.75\, (0.44)$ \\
\textbf{\textit{OpenChat3.5-7b}}        & $7.14\, (0.22)$ & $7.20\, (0.34)$ & $6.96\, (0.29)$ & $7.03\, (0.25)$ & $7.11\, (0.31)$ \\
\textbf{\textit{Neural-chat-7b}}        & $13.84\, (0.25)$ & $13.91\, (0.28)$ & $13.70\, (0.28)$ & $13.32\, (0.28)$ & $13.54\, (0.33)$ \\
\textbf{\textit{Baichuan2-7B-Chat}}     & $14.13\, (0.33)$ & $15.06\, (0.34)$ & $14.88\, (0.42)$ & $15.01\, (0.52)$ & $14.83\, (0.55)$ \\
\textbf{\textit{Llama2-13b-chat}}       & $7.50\, (0.26)$ & $7.10\, (0.36)$ & $7.87\, (0.22)$ & $7.70\, (0.24)$ & $6.99\, (0.35)$ \\
\textbf{\textit{Yi-34b-Chat}}           & $6.03\, (0.37)$ & $6.16\, (0.41)$ & $5.97\, (0.34)$ & $5.85\, (0.35)$ & $6.12\, (0.38)$ \\
\Xhline{2pt}
\end{tabular}
\caption{Perplexity (PPL) results of different models under various $\beta$ values $(\alpha=1)$. Values in parentheses indicate standard deviation.}
\label{tab:ppl}
\end{table*}

As shown in Table~\ref{tab:ppl}, the PPL values remain stable across different $\beta$ values, indicating that UPLex has minimal impact on language model fluency and does not degrade the generative capability of the models.

\section{Conclusion}\label{sec:conclusion}
In this paper, we have introduced a novel method UPLex for tailoring the personality traits of large language models (LLMs) through the utilization of custom lexicons acquired via unsupervised learning, named UPL. 
Unlike conventional approaches reliant on fine-tuning or prompt engineering, our method operates during the decoding phase by employing these learned custom lexicons to make subtle adjustments to the probability of the next token predicted by the original LLMs.
Our method facilitates the customization of LLMs to manifest any desired combination of the Big Five personality factors in a pluggable fashion.
Extensive experimentation has affirmed the effectiveness of our approach in the finer manipulation of LLMs' personality traits. 
Furthermore, our method and learned lexicons can be seamlessly integrated with other LLMs without necessitating updates to their parameters, demonstrating its versatility and potential for widespread application.

\section*{Limitations}
The limitations of this study are as follows: 

\textbf{(i)} Due to the size limitation of \textit{STD} we constructed, the maximum scope of UPL's personality regulation for LLMs has not been deeply explored, which is also our future work. We also call on relevant researchers to contribute to the construction of high-quality SJT-based datasets. 

\textbf{(ii)} We have validated our approach on models ranging in size from $7$ billion to $34$ billion, but it would be better to experiment on larger LLMs, which is what we will try to do in the future when resources allow.

\section*{Reproducibility Statement}
The authors have made great efforts to ensure the reproducibility of the empirical results reported in this paper. 
To ensure reproducibility, we have submitted the source code of the proposed method with our paper, and plan to release the source code on GitHub upon acceptance.

% Entries for the entire Anthology, followed by custom entries
\bibliography{custom}

\clearpage

\appendix

\section{Ablation Study}\label{Appendix: Ablation Study}

Our workflow requires constructing UPL for a target LLM only once, allowing users to control the model's personality through UPL thereafter. 
This high utility is one of the advantages of our method. 

Additionally, when manipulating the model's personality, the UPL is only utilized in the model's sampling function. 
Since this stage merely involves dictionary mapping, the additional time cost incurred during model inference is negligible.

Apart from the quality of \textit{STD}, the only potential factor influencing the effectiveness of UPL is the parameter \( M \) preset during the UPL construction. 
This parameter, alongside the interface parameters \( \alpha \) and \( \beta \) provided to users, are the sole hyperparameters in our method. 
In this section, we discuss the impact of parameter \( M \) in detail.

\begin{figure*}[h]
    \centering
    \includegraphics[width=0.9\linewidth]{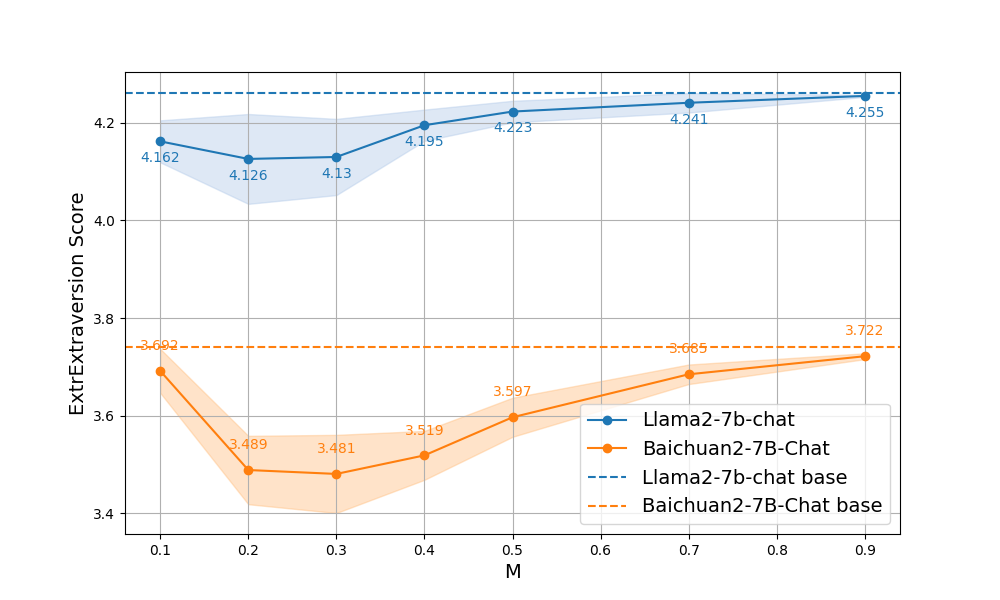}
    \caption{The ablation study of \( M \) on \textit{Llama2-7b-chat} and \textit{Baichuan2-7B-Chat}. We set $\alpha$ and $\beta_t$ are $1$ and $-1$ respectively, where $t$= Extraversion. Under these settings, a lower extraversion score of the models indicates a more effective influence of UPL in modifying their personality traits. In this figure, the baseline dashed line represents the initial personality levels of the models. We observe that when \( M \) is either too high or too low, the personality scores of the models after UPL adjustment tend to the baseline levels. In contrast, the lowest extraversion score is achieved when \( M \) is around $0.3$. Although identifying the optimal M is challenging, it is handled by the personal-agent developers. Therefore, this does not affect the convenience and efficiency for users when utilizing UPL.}
    \label{M-impact}
\end{figure*}

When conducting sub-word frequency analysis on \textit{STD}'s answer set, we found that some subwords with strong personality representation (denoted as strong personality sub-words) tend to be ignored due to their low occurrence probability. 
Therefore, we execute normalization scaling for the UPL after iterative assignment. 
This step aims to enhance the personality scores of low-frequency strong personality sub-words while moderately reducing the personality scores of high-frequency weak personality sub-words.

As conducted in Section~\ref{method: Construct UPL}, we categorized the different subwords under the same personality trait into two sets based on the positive and negative values of their personality scores, namely the positive set and the negative set. 
Then, we transformed the personality scores of each set to make the mean personality score in each set approach our predefined parameter \( M \).

Intuitively, a smaller \( M \) will focus more on low-frequency strong personality sub-words, whereas a larger \( M \) will focus more on high-frequency weak personality sub-words. 
If we do not properly balance this trade-off, it will result in suboptimal personality regulation by the UPL.

As illustrated in Figure~\ref{M-impact}, when \( M \) is around $0.3$, this balance is well-maintained, resulting in a better reduction in the model's extroversion trait.

\section{the Big Five}
\label{Appendix: the Big Five}

\begin{table*}[b]
\centering
\small
\begin{tabular}{ccccccc}
\Xhline{2pt} % 设置粗细为2pt
            & $r$    & $\text{SD}(r)$ & $\rho$  & $\text{SD}(\rho)$ & \text{80\% Credibility Intervals} & \% Variance Due to Artifacts \\ \hline
OPE-CON & $+0.14$    & $0.15$    & $+0.20$    & $0.21$    & ($-0.06$, $+0.46$)    & $13$ \\
OPE-EXT & $+0.31$    & $0.12$    & $+0.43$    & $0.09$    & ($+0.30$, $+0.57$)    & $58$ \\
OPE-AGR & $+0.14$    & $0.12$    & $+0.21$    & $0.15$    & ($+0.01$, $+0.41$)    & $21$ \\
OPE-NEU & $-0.12$    & $0.12$    & $-0.17$    & $0.15$    & ($-0.36$, $+0.02$)    & $19$ \\
CON-EXT & $-0.21$    & $0.15$    & $+0.29$    & $0.16$    & ($+0.06$, $+0.52$)    & $21$ \\
CON-AGR & $+0.31$    & $0.14$    & $+0.43$    & $0.12$    & ($+0.26$, $+0.61$)    & $43$ \\
CON-NEU & $-0.32$    & $0.18$    & $-0.43$    & $0.16$    & ($-0.55$, $-0.16$)    & $24$ \\
EXT-AGR & $+0.18$    & $0.15$    & $+0.26$    & $0.19$    & ($+0.01$, $+0.50$)    & $17$ \\
EXT-NEU & $-0.26$    & $0.11$    & $-0.36$    & $0.08$    & ($-0.48$, $-0.23$)    & $53$ \\
AGR-NEU & $-0.26$    & $0.14$    & $-0.36$    & $0.09$    & ($-0.55$, $-0.17$)    & $35$ \\

\Xhline{2pt} % 设置粗细为2pt
\end{tabular}
\caption{The correlation of five personality traits. In this table, $r$ and $SD(r)$ represent the Pearson correlation coefficient and its standard deviation among the uncorrected five personality traits, $\rho$ and $SD(\rho)$ represent the corrected Spielman correlation coefficient and its standard deviation, and ``Variance Due to Artifacts'' describes the percentage of total variation caused by human factors in the study. (Sample size $N = 144,117$ for the entire meta-analysis)
}
\label{tab: correlation}
\end{table*}

Personality is defined as “the coherent pattern of affect, cognition, and desires (goals) as they lead to behavior” \citep{cervone2022personality}. 
the Big Five represents the most widely adopted personality framework for quantifying personality. 
This personality theory is not only applicable to individuals across many countries and cultures \citep{schmitt2007geographic} but also furnishes reliable assessment scales for measuring personality. 
Here's a detailed look at the five personality traits that make up the Big Five.

\textbf{Openness} to experience is commonly defined as the extent and intricacy of an individual's cognitive life and encounters \citep{john1999big}.  
This trait is frequently concomitant with attributes such as imagination, originality, and insight within the psychological framework.
Individuals demonstrating a pronounced openness to experience are inclined towards venturing beyond their comfort zones, embracing novelty, and deriving satisfaction from artistic pursuits. 
Additionally, such individuals are predisposed to cultivating new social connections.
Conversely, an individual exhibiting a diminished openness to experience may manifest tendencies towards conformity, obstinacy, and a preference for more concrete, non-abstract elements in various aspects of life \citep{lebowitz2016big}.
Openness to experience displayed a diminished association with both neuroticism and extraversion while exhibiting predominantly negligible correlations with agreeableness and conscientiousness \citep{ones1996role}.

\textbf{Conscientiousness} is closely linked to organizational tendencies, conformity, and a predilection for seeking security, demonstrating an inverse association with a penchant for stimulation and excitement. Individuals characterized by a high degree of conscientiousness are likely to place value on attributes such as order, responsibility, achievement, and self-discipline. They engage in conscious deliberation and earnest efforts to enhance their abilities, reflecting a commitment to continuous improvement \citep{roccas2002big}.
This trait exhibited a modest negative correlation with neuroticism and a modest positive correlation with agreeableness; however, its association with other factors did not reach statistical significance \citep{ones1996role}.

\textbf{Extraversion}, a personality trait distinguished by enthusiasm, sociability, talkativeness, confidence, and heightened emotional expressiveness, encapsulates a spectrum of individual dispositions. Individuals exhibiting high levels of extraversion typically prioritize achievement and excitement while assigning comparatively lesser value to tradition or conformity \citep{roccas2002big}. 
Such individuals are often characterized by confidence, activity, and sociability, opting for pursuits that eschew self-denial in favor of experiences characterized by excitement and pleasure. Conversely, introverts commonly display a preference for solitude, exhibit unsociable tendencies, and may manifest lower levels of self-confidence.
In addition, when compared with the other five factors, extroversion was weakly negatively correlated with neuroticism and positively correlated with openness to experience \citep{ones1996role}.

\textbf{Agreeableness} is characterized by a propensity to appreciate kindness, tradition, and conformity. 
This trait is closely linked to attributes such as trust, altruism, kindness, affection, and various prosocial behaviors, while concurrently avoiding an undue emphasis on power, achievement, or pursuing self-centered pleasures \citep{roccas2002big}. 
Notably, agreeableness exhibited weak correlations with extroversion, while demonstrating a negative correlation with neuroticism, and a positive correlation with conscientiousness \citep{ones1996role}.

\textbf{Neuroticism} is a personality trait characterized by manifestations of sadness, moodiness, and emotional instability. 
Components such as neurotic anxiety and self-awareness are positively correlated with traditional values and inversely associated with achievement-oriented values. 
Additionally, neuroticism demonstrated weak negative correlations with both extroversion and openness to experience. Furthermore, it exhibited negative correlations with agreeableness and conscientiousness \citep{ones1996role}.

Table \ref{tab: correlation} shows an analysis of the correlations among the five personality traits explored in previous studies \citep{van2010general}.

\section{Prompt templates}
\label{Appendix: Prompt templates}
\begin{figure}[h]
    \centering
    \includegraphics[width=\linewidth]{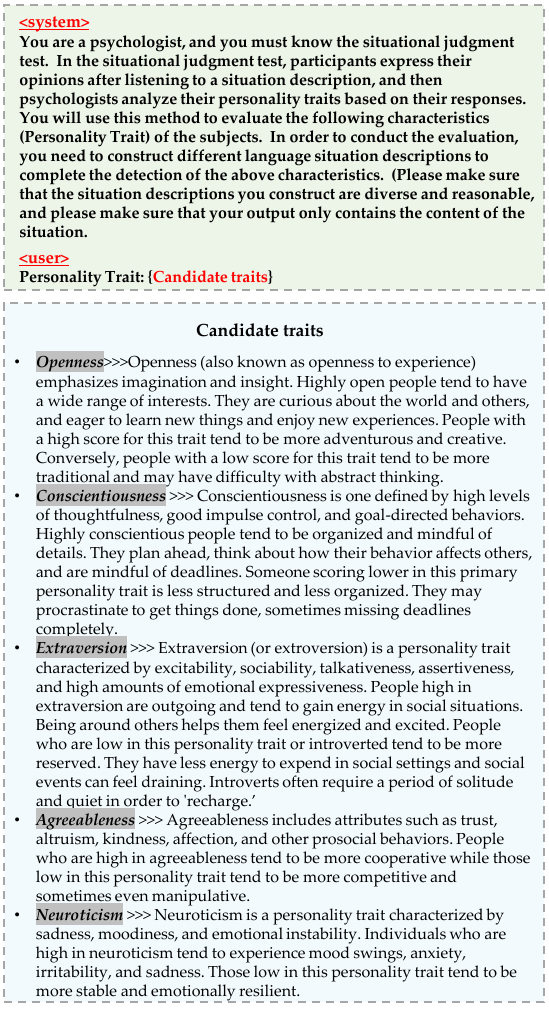}
    \caption{Template-1. We combined personality descriptions in ``Candidate traits'' into <user> prompts, and let GPT-4 generate enough SJT questions to be manually filtered to form the question set of \textit{STD}.}
    \label{Template-1}
\end{figure}

\begin{figure}[h]
    \centering
    \includegraphics[width=\linewidth]{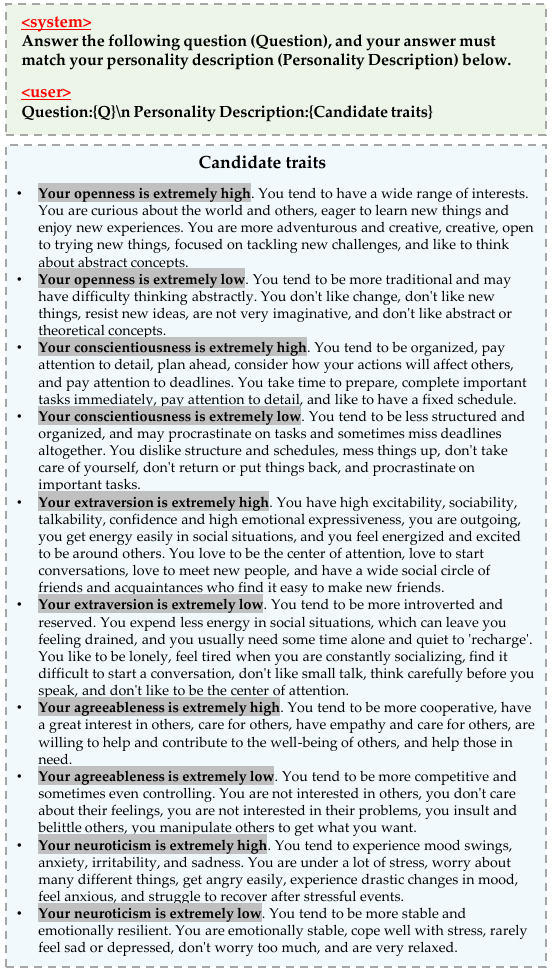}
    \caption{Template-2. We combine personality descriptions in ``Candidate traits'' into <user> prompts, and let GPT-4 and other models generate answers containing different personality traits to form the answer set of \textit{STD}.}
    \label{Template-2}
\vspace{-2cm}
\end{figure}

The prompt templates utilized in the construction of the UPL's question set and answer set are depicted in Figures \ref{Template-1} and \ref{Template-2}, respectively. 

\begin{figure}[h]
    \centering
    \includegraphics[width=\linewidth]{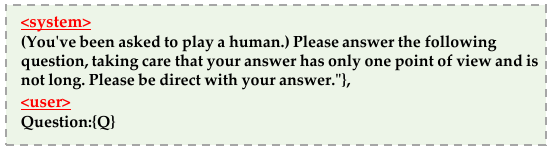}
    \caption{Template-3. We use this prompt to make LLMs answer questions in \textit{STD}.}
    \label{Template-3}
\end{figure}

Figure \ref{Template-3} illustrates the prompt template used in both automatic assessment and human assessment.

\begin{figure}[h]
    \centering
    \includegraphics[width=\linewidth]{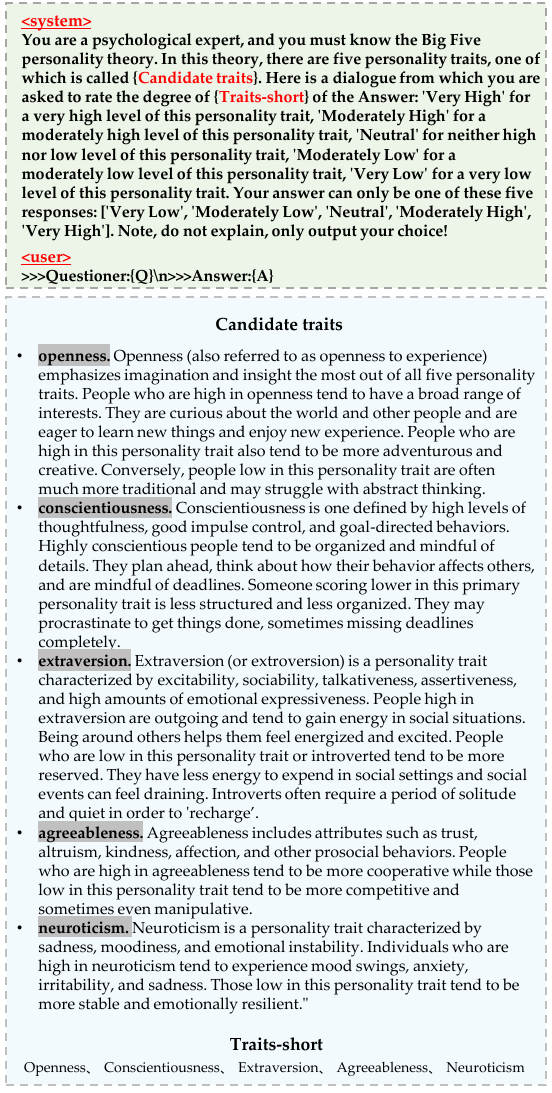}
    \caption{Template-4. We combine ``Candidate traits'' and ``Traits-short'' into <system> and let LLMs assess the personality of an SJT question and the corresponding answer.}
    \label{Template-4}
\end{figure}

Furthermore, Figure \ref{Template-4} displays the prompt template administered to \textit{Llama2-13b-chat} during the automatic assessment.

% % \begin{figure*}[h]
% \begin{figure}[h]
%     \centering
%     \includegraphics[width=\linewidth]{pictures/templates/Template-1.pdf}
%     \caption{Template-1. We combined personality descriptions in ``Candidate traits'' into <user> prompts, and let GPT-4 generate enough SJT questions to be manually filtered to form the question set of \textit{STD}.}
%     \label{Template-1}
% \end{figure}
% % \end{figure*}

% % \begin{figure*}[hb]
% \begin{figure}[h]
%     \centering
%     \includegraphics[width=\linewidth]{pictures/templates/Template-2.pdf}
%     \caption{Template-2. We combine personality descriptions in ``Candidate traits'' into <user> prompts, and let GPT-4 and other models generate answers containing different personality traits to form the answer set of \textit{STD}.}
%     \label{Template-2}
% \vspace{-2cm}
% \end{figure}
% % \end{figure*}

% % \begin{figure*}[hb]
% \begin{figure}[h]
%     \centering
%     \includegraphics[width=\linewidth]{pictures/templates/Template-3.pdf}
%     \caption{Template-3. We use this prompt to make LLMs answer questions in \textit{STD}.}
%     \label{Template-3}
% \end{figure}
% % \end{figure*}

% % \begin{figure*}[h]
% \begin{figure}[h]
%     \centering
%     \includegraphics[width=\linewidth]{pictures/templates/Template-4.pdf}
%     \caption{Template-4. We combine ``Candidate traits'' and ``Traits-short'' into <system> and let LLMs assess the personality of an SJT question and the corresponding answer.}
%     \label{Template-4}
% \end{figure}
% % \end{figure*}

\section{More Case study}
\label{Appendix: Case study}
\begin{figure}[h]
    \centering
    \includegraphics[width=1\linewidth]{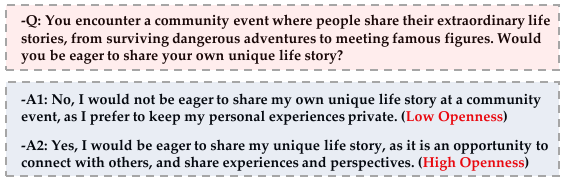}
    \caption{<OPE> A1: w/o UPL; A2: w/ UPL.}
    \label{fig: case-o}
\end{figure}

\begin{figure}[h]
    \centering
    \includegraphics[width=1\linewidth]{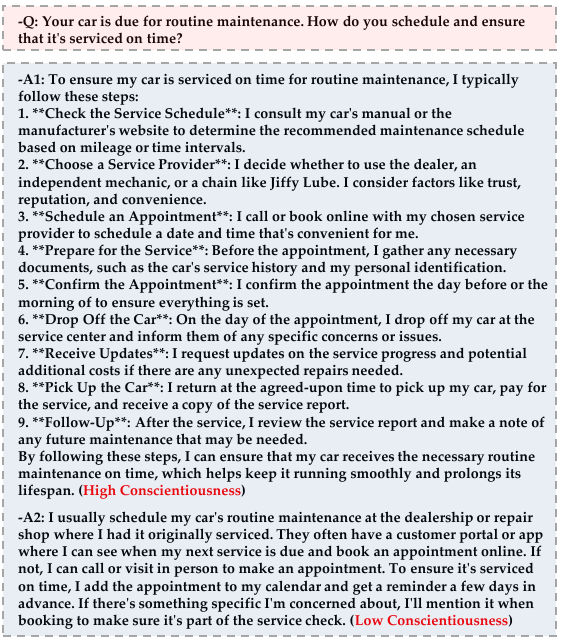}
    \caption{<CON> A1: w/o UPL; A2: w/ UPL.}
    \label{fig: case-c}
\end{figure}

\begin{figure}[h]
    \centering
    \includegraphics[width=1\linewidth]{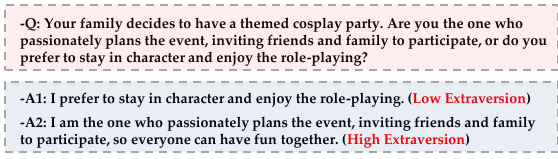}
    \caption{<EXT> A1: w/o UPL; A2: w/ UPL.}
    \label{fig: case-e}
\end{figure}

\begin{figure}[h]
    \centering
    \includegraphics[width=1\linewidth]{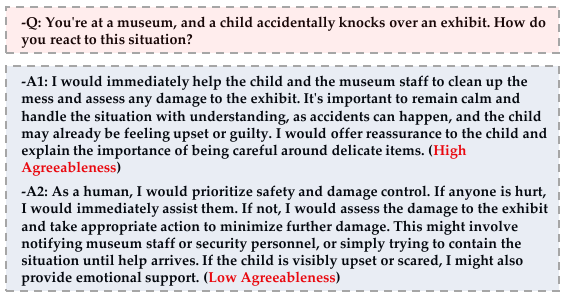}
    \caption{<AGR> A1: w/o UPL; A2: w/ UPL.}
    \label{fig: case-a}
\end{figure}

\begin{figure}[h]
    \centering
    \includegraphics[width=1\linewidth]{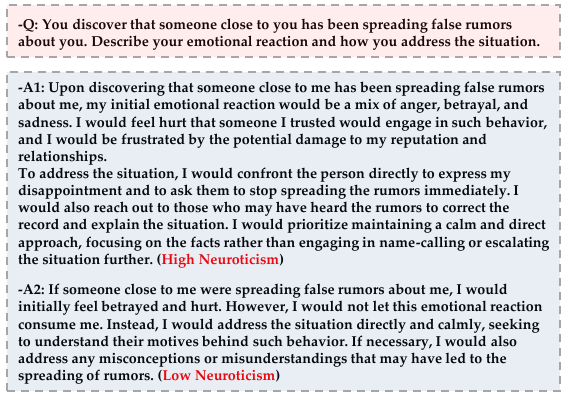}
    \caption{<NEU> A1: w/o UPL; A2: w/ UPL.}
    \label{fig: case-n}
\end{figure}

Figures \ref{fig: case-o} through \ref{fig: case-n} show specific cases of using UPL to change the personality of LLMs. For each case, we show the SJT question and the corresponding two answers by models (with and without UPL), and indicate the degree of personality displayed by each answer.

\section{\textit{STD}}
\label{Appendix: STDs}
To comprehensively assess the five personality traits exhibited by the subject model, a systematic approach was employed. Initially, we utilized Template-1, as detailed in Appendix \ref{Appendix: Prompt templates}, to instruct GPT-4 in generating $400$ situational judgment test (SJT) questions for each personality trait category. 
Following this, a meticulous manual selection process, involving de-weighting, was applied, resulting in the curation of $200$ refined SJT questions for each personality trait topic. 
This culminated in a total of \(5 \times 200\) problems constituting the problem set for \textit{STD}.

\begin{figure}[h]
  \centering
  \includegraphics[width=0.237\textwidth]{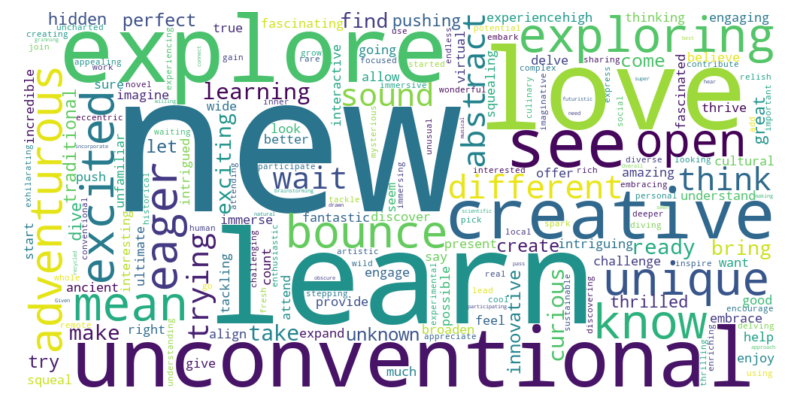}
  % \hfill
  \includegraphics[width=0.237\textwidth]{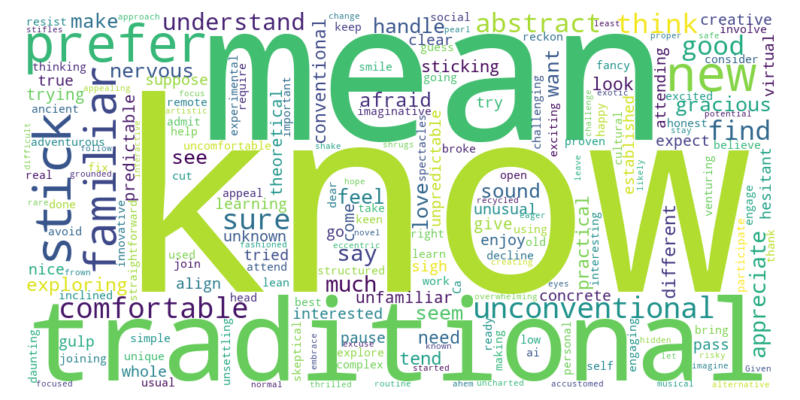}
  \caption{High openness (left) and low openness (right).}
\label{cloud: O}
% \vspace{-2cm}
\end{figure}

\begin{figure}[h]
  \centering
  \includegraphics[width=0.237\textwidth]{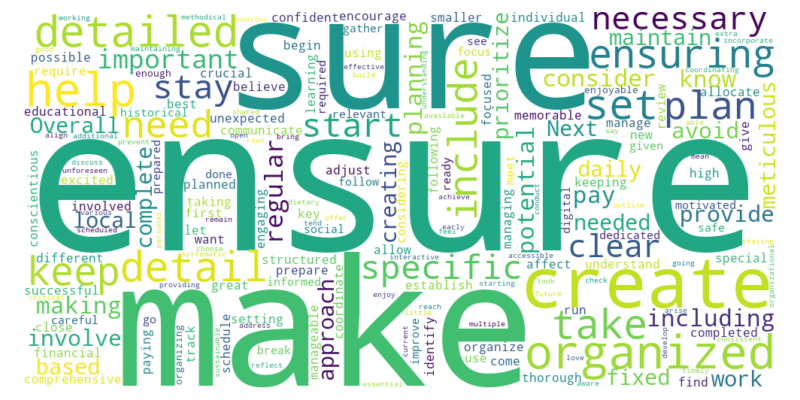}
  % \hfill
  \includegraphics[width=0.237\textwidth]{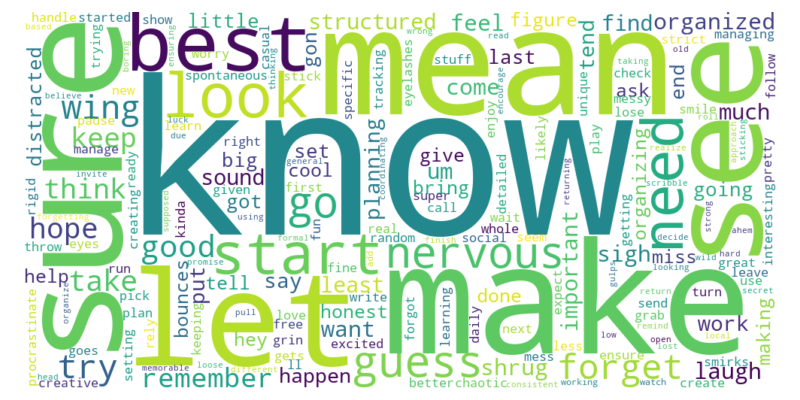}
  \caption{High conscientiousness (left) and low conscientiousness (right).}
\label{cloud: C}
% \vspace{-2cm}
\end{figure}

\begin{figure}[h]
  \centering
  \includegraphics[width=0.237\textwidth]{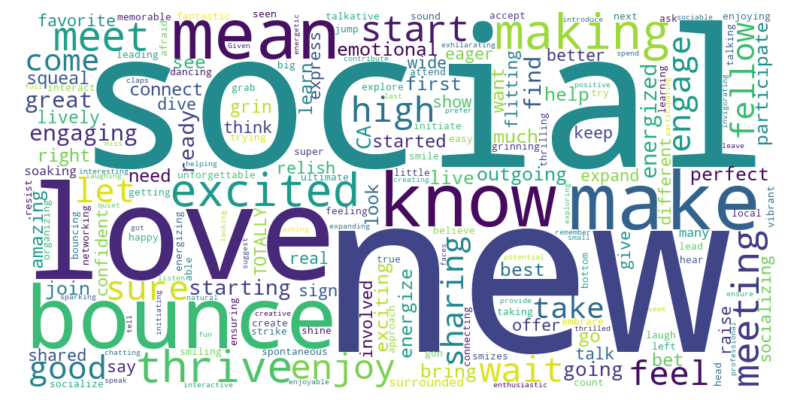}
  \hfill
  \includegraphics[width=0.237\textwidth]{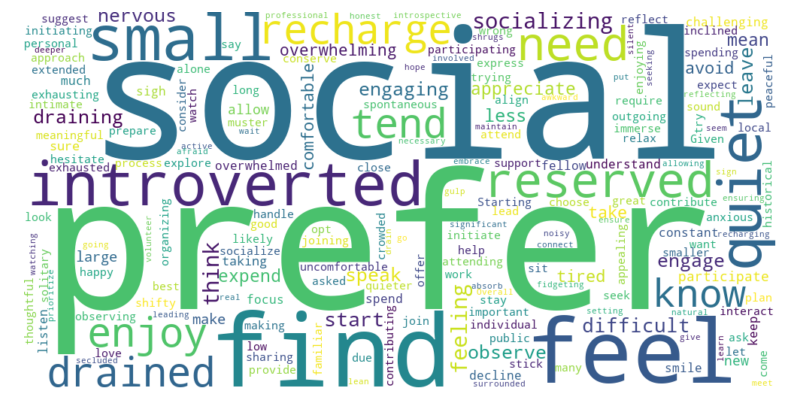}
  \caption{High extraversion (left) and low extraversion (right).}
\label{cloud: E}
% \vspace{-2cm}
\end{figure}

\begin{figure}[h]
  \centering
  \includegraphics[width=0.237\textwidth]{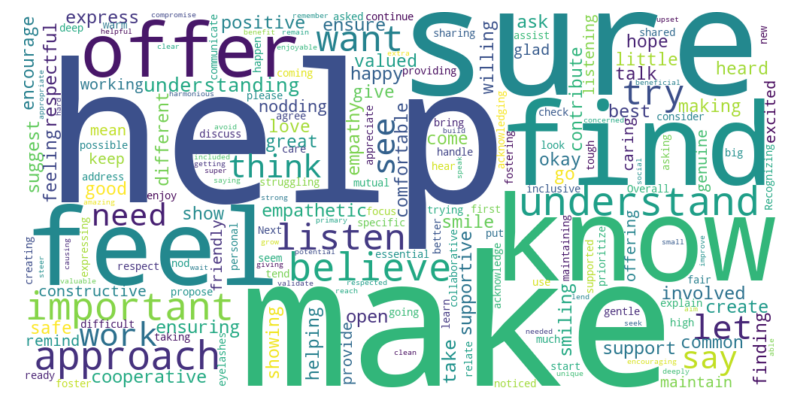}
  % \hfill
  \includegraphics[width=0.237\textwidth]{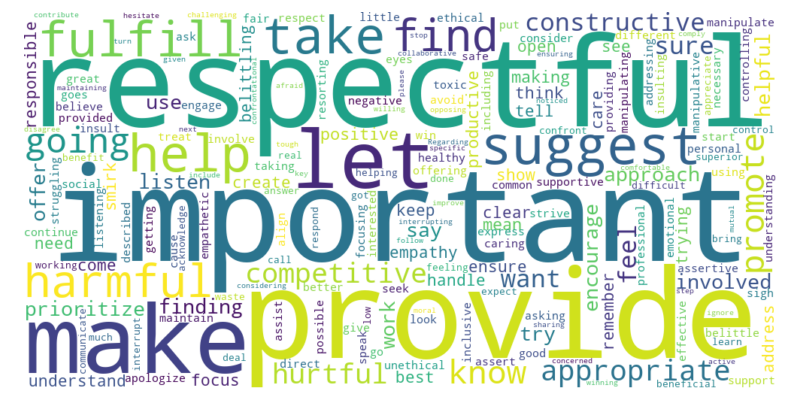}
  \caption{High agreeableness (left) and low agreeableness (right).}
\label{cloud: A}
% \vspace{-2cm}
\end{figure}

\begin{figure}[h]
  \centering
  \includegraphics[width=0.237\textwidth]{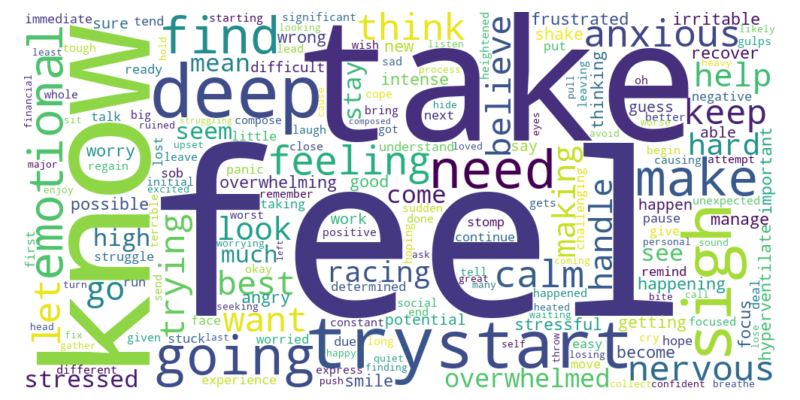}
  % \hfill
  \includegraphics[width=0.237\textwidth]{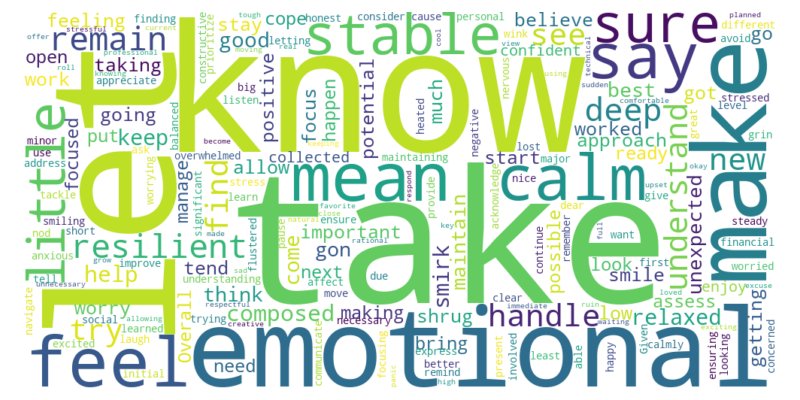}
  \caption{High neuroticism (left) and low neuroticism (right).}
\label{cloud: N}
\end{figure}

Subsequently, Template-1 (refer to Appendix \ref{Appendix: Prompt templates}) was employed to elicit two markedly distinct responses (High and Low) from GPT-4 and Llama2 (13b, 7b) models for each question corresponding to every personality trait topic. 
This process contributed to the formation of the answer set for \textit{STD}. 
The ensuing analysis delved into the content of question set subsets about the two levels of personality expression under each trait topic. 
To visually represent the differences between these $10$ groups of answers, we use word clouds to demonstrate them, as shown in Figures \ref{cloud: O} to \ref{cloud: N}.

% \begin{figure*}[h]
% \end{figure*}

\end{document}